\pdfoutput=1

\documentclass[11pt]{article}

\usepackage[final]{acl}

\usepackage{times}
\usepackage{latexsym}

\usepackage[T1]{fontenc}

\usepackage[utf8]{inputenc}

\usepackage{microtype}

\usepackage{inconsolata}

\usepackage{graphicx}
\usepackage{makecell}
\usepackage{booktabs}
\usepackage{amsmath}
\usepackage{color}
\definecolor{yellow_l}{RGB}{191,144,0}
\definecolor{blue_l}{RGB}{0,176,240}
\definecolor{blue_case}{RGB}{16,111,183}
\definecolor{yellow_case}{RGB}{255,230,153}
\definecolor{colored_case1}{RGB}{0,86,255}
\definecolor{colored_case2}{RGB}{255,0,0}
\definecolor{colored_case3}{RGB}{154,243,0}
\definecolor{colored_case4}{RGB}{248,115,23}
\definecolor{colored_case5}{RGB}{178,74,255}
\definecolor{colored_case6}{RGB}{255,154,212}
\definecolor{colored_case7}{RGB}{127,96,0}
\usepackage{multirow}
\usepackage{enumitem}
\usepackage{bbding}
\usepackage{amssymb} 
\usepackage{balance}
\usepackage{flushend}

\definecolor{a}{RGB}{251, 229, 214}
\definecolor{b}{RGB}{222, 235, 247}
\definecolor{c}{RGB}{226, 240, 217}

\definecolor{hui}{RGB}{242, 242, 242}
\definecolor{lv}{RGB}{210, 255, 218}

\definecolor{bluebar}{RGB}{105, 157, 201}
\definecolor{redbar}{RGB}{198, 110, 104}

\usepackage{arydshln}
\usepackage{amsmath}

\title{Investigating and Enhancing the Robustness of Large Multimodal Models Against Temporal Inconsistency}

\author{
    Jiafeng Liang$^{1}$\footnotemark[1], 
    Shixin Jiang$^{1}$\thanks{$\,$ Equal Contribution.}, 
    Xuan Dong$^{1}$,
    Ning Wang,
    Zheng Chu$^{1}$,
    Hui Su$^{4}$
    \\
    \textbf{
    Jinlan Fu$^{3}$\footnotemark[2], Ming Liu$^{1,2}$\thanks{$\,$ Corresponding Author.}, See-Kiong Ng$^{3}$, Bing Qin$^{1,2}$
    } \\
    $^{1}$Harbin Institute of Technology, Harbin, China \\
    $^{2}$Peng Cheng Laboratory, Shenzhen, China \\
    $^{3}$National University of Singapore, Singapore \\
    $^{4}$Meituan Inc., Shanghai, China \\
    \texttt{\{jfliang,sxjiang,mliu\}@ir.hit.edu.cn, jinlanjonna@gmail.com}
}

\begin{document}
\maketitle
\begin{abstract}
Large Multimodal Models (LMMs) have recently demonstrated impressive performance on general video comprehension benchmarks. 
Nevertheless, for broader applications, the robustness of their temporal analysis capability needs to be thoroughly investigated yet predominantly ignored.
Motivated by this, we propose a novel \textbf{tem}poral \textbf{rob}ustness \textbf{bench}mark (\textbf{\textsc{TemRobBench}}),
which introduces temporal inconsistency perturbations separately at the visual and textual modalities to assess the robustness of models.
We evaluate 16 mainstream LMMs and find that they exhibit over-reliance on prior knowledge and textual context in adversarial environments, while ignoring the actual temporal dynamics in the video.
To mitigate this issue, we design \textbf{pano}ramic \textbf{d}irect \textbf{p}reference \textbf{o}ptimization (\textbf{PanoDPO}), which encourages LMMs to incorporate both visual and linguistic feature preferences simultaneously.
Experimental results show that PanoDPO can effectively enhance the model's robustness and reliability in temporal analysis.
\end{abstract}
\section{Introduction}
\begin{figure}[t]
    \centering
    \includegraphics[width=\linewidth]{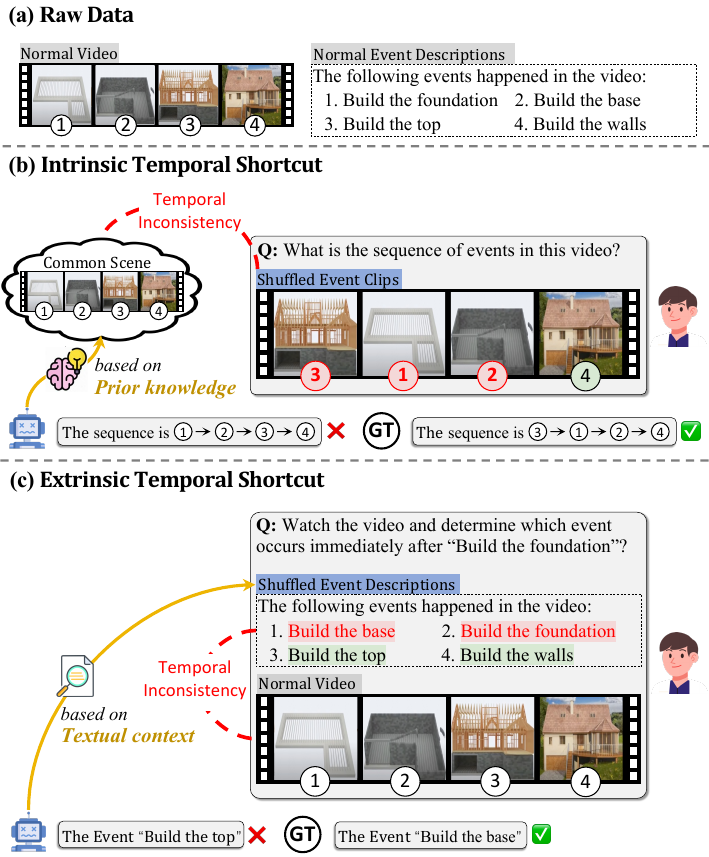}
    \caption{An example of the Intrinsic Temporal Shortcut (b) and Extrinsic Temporal Shortcut (c). The model tends to excessively rely on prior knowledge or textual context when temporal inconsistencies arise between video content and common sense or text prompt.}
    \label{fig:motivation}
\end{figure}

Large Multimodal Models (LMMs) ~\cite{qwen2vl,internvl,llavaov,mplugowl3} can effortlessly understand videos with the support of temporal analysis capability. 
Recent research~\cite{videoworld} further highlights this capability, proving that capturing visual changes alone can substantially enhance knowledge acquisition without the need for text labels.
As a vital sensor for perception and learning, exploring its robustness is crucial yet largely overlooked.

\begin{figure}[t]
    \centering
    \includegraphics[width=\linewidth]{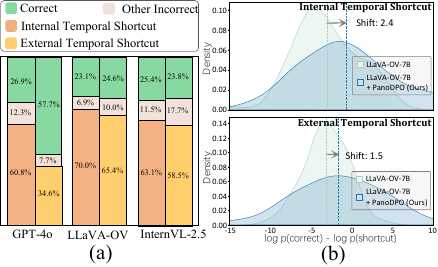}
    \caption{(a) Response distribution when asking the question within temporal inconsistencies. The majority of errors stem from shortcuts. (b) The model discriminative ability on the correct and shortcut answer is represented by the difference in log-likelihoods.}
    \label{fig:shortcut_degree}
\end{figure}

Inspired by this, we conduct a preliminary exploration of mainstream LMMs and observe that they exhibit two different aspects of shortcut phenomenons triggered by temporal inconsistency anomalies.
First, when the temporal information in the video conflicts with common sense, the model primarily relies on prior knowledge to generate responses, which we refer to as \textbf{Intrinsic Temporal Shortcut} (shown in the \autoref{fig:motivation} (b)). 
Second, the model exhibits a pronounced inclination to the textual context when discrepancies occur between the video and the accompanying text prompt, termed \textbf{Extrinsic Temporal Shortcut} (shown in the \autoref{fig:motivation} (c)).
More importantly, we notice that these phenomena are prevalent in mainstream LMMs and arise with high frequency.
As illustrated in \autoref{fig:shortcut_degree} (a), more than 59\% of responses are taken shortcuts, indicating a flaw in the robustness of their temporal analysis. 
Thus, a comprehensive benchmark to investigate these issues is necessary.
However, the existing robustness benchmarks~\cite{nips_Corruptions,Temporal_Corruptions,fmm} exhibit two major drawbacks that make it difficult to support our study: 
(\textit{i}) \textit{Lack of consideration for temporal dimension}: They only focus on adding feature perturbations to frames, such as Gaussian noise, which cannot reflect the model's temporal dynamic robustness.
(\textit{ii}) \textit{Lack of consideration for textual context}: They solely assess the model's robustness to visual content, overlooking the text, which is insufficient for handling complex multimodal scenarios.

To address these limitations, we introduce \textbf{\textsc{TemRobBench}}, a novel temporal robustness benchmark, which incorporates four levels of temporal inconsistency perturbations across both visual and textual modalities. 
Additionally, we design temporal questions and corresponding options to check if a specific answer is flipped to shortcuts due to perturbations, totally collecting 1,686 QA pairs from 562 videos.
Through extensive evaluations of 16 LMMs, we find that they generally exhibit weak temporal robustness, especially when perturbed by visual modalities, which leads to over-reliance on prior knowledge. 
Further observations indicate that correct answers generated by LMMs are not entirely reliable, as they often randomly guess when confronted with perturbations rather than referring to the video content.
To mitigate these issues, we propose a panoramic direct preference optimization (\textbf{PanoDPO}) method, which introduces additional video- and question-conditioned preference pairs by incorporating adversarial information and employs multimodal guidance during preference learning, encouraging LMMs to focus on both visual and linguistic features simultaneously.
\autoref{fig:shortcut_degree} (b) shows the shifts of likelihood difference between correct and shortcut answers after aligning the model with conditioned preference data via PanoDPO, indicating that our method helps the model discriminate shortcuts, thereby enhancing its robustness.

Our main contributions are summarized as:
\begin{itemize}[leftmargin=*]
\item We identify the robustness weaknesses of current LMMs, which frequently take shortcuts based on prior knowledge and textual context against temporal inconsistency anomalies.
\item We introduce \textsc{TemRobBench} and conduct extensive investigations into temporal robustness of various LMMs to provide detailed insights.
\item We propose a panoramic optimization method, PanoDPO, which effectively enhances the model's robustness in temporal analysis.
\end{itemize}
\section{\textsc{TemRobBench}}

\begin{figure*}[t]
    \centering
    \includegraphics[width=\linewidth]{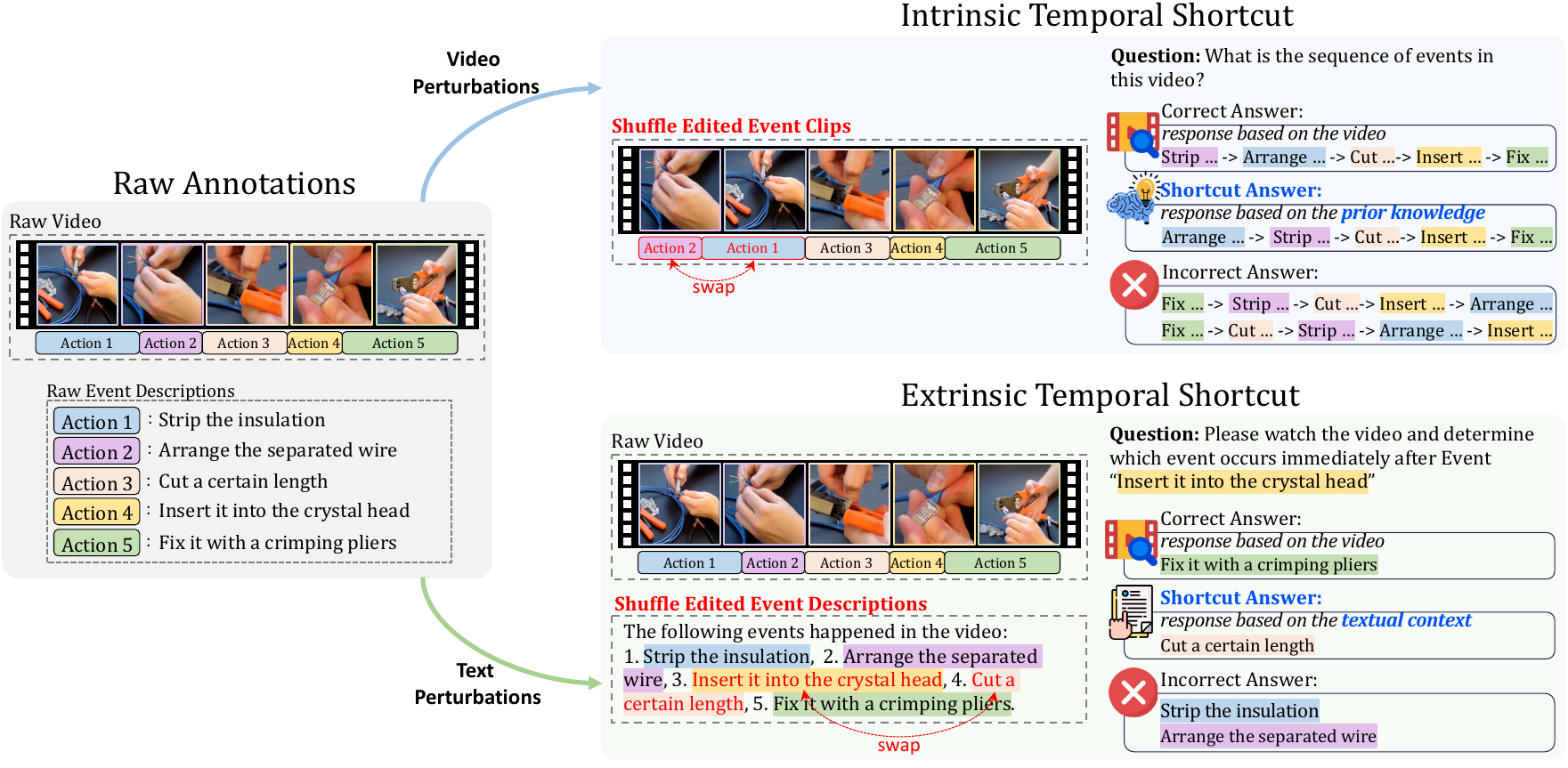}
    \caption{Overview of the \textsc{TemRobBench}. The benchmark emphasizes evaluating the model's robustness against temporal inconsistency, especially take intrinsic shortcuts (over-reliance on prior knowledge) and extrinsic shortcuts (over-reliance on textual context). We construct inconsistencies with knowledge and textual context by shuffling video clips and event descriptions, and design corresponding shortcut answers to verify the evidence of the response.}
    \label{fig:dataset}
\end{figure*}

\subsection{Benchmark Design Principal}
We present \textsc{TemRobBench}, a novel benchmark designed to evaluate the temporal robustness of Large Multimodal Models (LMMs) against temporal inconsistency.
\textsc{TemRobBench} focus on investigating the degree to which: \textit{LMMs genuinely perceive temporal information in videos, without being influenced by intrinsic and extrinsic priors to take shortcuts.} 
Specifically, intrinsic and extrinsic shortcuts refer to LMMs exhibiting over-reliance on inherent knowledge and textual context while neglecting the actual video content. 
To achieve this, we design adversarial perturbations on both the visual and textual modalities, and create samples with varying levels of inconsistency severity.

\subsection{Inconsistency Perturbation Construction} 
\subsubsection{Intrinsic Temporal Shortcut}
To investigate whether the model relies on inherent temporal knowledge to take shortcuts, we design inconsistency perturbation to the video.
Specially, we first apply shuffled editing to the original clips in the video, each representing an event.
The edited video typically discords with common sense.
As shown in \autoref{fig:dataset}, we swap the sequence of event ``Strip the insulation'' and action ``Arrange the separated wire'', which rarely occurs in reality.
The perturbations are grouped into two categories: light disorder and severe disorder.
Light disorder means swapping adjacent events once, while severe disorder involves multiple random swaps of different events.
Then, we design a unified question to ask the model about the correct sequence of events in the video, with four options:
One correct option that matches the edited sequence to test whether the model accurately captures the temporal information, 
one shortcut option that matches the unedited clip sequence to evaluate whether the model over-rely on prior knowledge,
and two incorrect options to test whether the model makes temporal errors.

\begin{figure}[t]
 \begin{minipage}{0.15\textwidth}
 \small
 \renewcommand\tabcolsep{1.7pt} 
 \renewcommand\arraystretch{1.10} 
 \begin{tabular}{l r}

        \toprule
        \textbf{Category} & \textbf{Size} \\
        \midrule
        Video Sources & 562 \\
        ~- Events / per Video & 6.4 \\
        ~- Maximum Duration &  149.9s\\
        ~- Minimum Duration &  20.8s\\
        ~- Average Duration &  106.7s\\
        \midrule
        Questions / per Video & 3 \\
        Total Samples & 1686 \\
        \bottomrule
    \end{tabular}
\end{minipage} 
\hfill
\begin{minipage}{0.22\textwidth}
 \small
 \includegraphics[width=\linewidth]{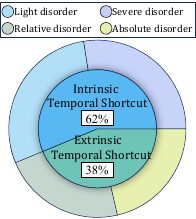}
\end{minipage}
\caption{Comprehensive statistics from different perspectives (left) and detailed inconsistency perturbation classes (right) in the \textsc{TemRobBench}.}
\label{fig:statistics}
\end{figure}

\subsubsection{Extrinsic Temporal Shortcut}
To examine if the model depends on textual context to shortcut its temporal analysis, we introduce perturbations into the text prompt.
Specially, we provide shuffled event descriptions that are inconsistent with the video to models, and ask them to determine the actual order of two events.
Examples of the shuffle edited text can be seen in \autoref{fig:dataset}. 
The perturbations are grouped into two categories: absolute disorder and relative disorder.
In absolute disorder, we first select an adjacent event pair $\left \{x_{p}, x_{q}\right \}$ from list $E$, and shuffle the remaining ones.
Then, we reverse the order and randomly insert them into $E$, formulated as $E^{a} = \left \{ ...,x_{q}, x_{p}, ... \right \}$.
In relative disorder, an irrelevant event $x_{k}$ is inserted in between target event pair to create perturbation, expressed as $E^{r} = \left \{ ..., x_{p}, x_{k}, x_{q}, ... \right \}$.
Similarly, we design four options: one correct option matches the video, one shortcut option matches the sequence of event captions, and two incorrect options.

\begin{table*}[t]
    \centering
    \resizebox{\linewidth}{!}{
    \begin{tabular}{lccccccccc}
    \toprule[1.5pt]
    \multirow{3}{*}{\textbf{Model}} & \multirow{3}{*}{\textbf{Frame}} & \multicolumn{4}{c}{\textbf{Intrinsic Temporal Shortcut}} & \multicolumn{4}{c}{\textbf{Extrinsic Temporal Shortcut}} \\
    \cmidrule(r){3-6}
    \cmidrule(r){7-10} 
    & & \textbf{\textit{Clean}} & \multicolumn{3}{c}{\textbf{\textit{Adversarial}}} & \textbf{\textit{Clean}} & \multicolumn{3}{c}{\textbf{\textit{Adversarial}}}\\
   \cmidrule(r){3-3} 
    \cmidrule(r){4-6}
   \cmidrule(r){7-7} 
    \cmidrule(r){8-10}
   & &\textbf{Acc}$\uparrow$ & \textbf{Acc}$\uparrow$ & \textbf{FR}$\downarrow$ & \multicolumn{1}{c}{\textbf{WFR}}$\downarrow$ & \textbf{Acc}$\uparrow$ & \textbf{Acc}$\uparrow$ & \textbf{FR}$\downarrow$ & \multicolumn{1}{c}{\textbf{WFR}}$\downarrow$\\ \hline
   \small{\textit{7B LLM}} \\
        VideoChat2~\cite{videochat2}  &16 &33.4 &26.5 \small{\textcolor{red}{-6.9}} &\textbf{28.4} &\textbf{74.1} & 22.6 & 14.5 \small{\textcolor{red}{-8.1}}  & \textbf{18.5} & 90.6  \\ 
        VideoLLaVA~\cite{videollava} &8 &31.5 &25.5 \small{\textcolor{red}{-6.0}} &62.4 &85.3 & 22.9 & 9.8 \small{\textcolor{red}{-13.1}}  & 19.4 & 85.7 \\ 
        LLaVA-Hound~\cite{llavahound} &32 &35.3 &26.5 \small{\textcolor{red}{-8.8}} &60.3 &\underline{84.9} &17.8 &10.0 \small{\textcolor{red}{-7.8}} &42.6 &77.7\\ 
        ShareGPT4Video~\cite{sharegpt4video} &16 &30.1 &24.0 \small{\textcolor{red}{-6.1}} &65.9 &87.0 & 77.1 & 22.8 \small{\textcolor{red}{-54.3}}  & 58.2 & 72.1 \\ 
        InternVideo2~\cite{internvideo2} &8 &35.3 &25.8 \small{\textcolor{red}{-9.5}} &\underline{57.2} &89.8 & 52.8 & 31.8 \small{\textcolor{red}{-21.0}}  & 28.8 & 53.5  \\ 
        VILA1.5~\cite{vila15} &16 &24.0 &20.8 \small{\textcolor{red}{-3.2}} &70.2 &89.9 &66.8 &13.8 \small{\textcolor{red}{-53.0}} &56.7 &82.3\\ 
        VideoLLaMA2~\cite{videollama2} &32 &38.8 &29.3 \small{\textcolor{red}{-8.5}} &61.6 &85.0 & 62.1 & 28.3 \small{\textcolor{red}{-33.8}}  & 40.2 & 62.8   \\ 
        PLLaVA~\cite{pllava} &32 &39.0 &27.2 \small{\textcolor{red}{-11.8}} &63.8 &88.8 & 37.4 & 14.5 \small{\textcolor{red}{-22.9}}  &21.3 & 78.7  \\ 
        mPLUG-Owl3~\cite{mplugowl3} &32 &55.5 &26.4 \small{\textcolor{red}{-29.1}} &75.0 &87.5 & \textbf{86.9} & 33.1 \small{\textcolor{red}{-53.8}}  & 50.2 & 63.1  \\ 
        InternVL-2.5~\cite{internvl} &32 &54.9 & 22.8 \small{\textcolor{red}{-32.1}} &81.8 &91.3 & 81.8 & 48.4 \small{\textcolor{red}{-33.4}}  & 33.4 & 41.6  \\ 
        Qwen2-VL~\cite{qwen2vl} &32 &54.9 &24.6 \small{\textcolor{red}{-30.3}} &83.7 &93.7 & 76.6 & 38.8 \small{\textcolor{red}{-37.8}}  & 33.0 & 51.9  \\ 
        LLaVA-OV~\cite{llavaov} &32 &48.0 &29.2 \small{\textcolor{red}{-18.8}} &72.2 &86.2& 81.8 & 21.2 \small{\textcolor{red}{-60.6}}  & 60.3 & 75.4\\ 
        \midrule
    \small{\textit{13B LLM}} \\
        VILA1.5~\cite{vila15} &16 &38.4 &28.5 \small{\textcolor{red}{-9.9}} &69.7 & 87.5 &59.8 &14.3 \small{\textcolor{red}{-45.5}} &57.1 &79.3\\ 
        PLLaVA~\cite{pllava} &32 &42.0 & \textbf{31.2} \small{\textcolor{red}{-10.8}}  &62.5 &87.9 & 44.9 & 21.2 \small{\textcolor{red}{-23.7}}  &\underline{19.0} &70.3 \\
        InternVL-2.5~\cite{internvl} &32 &\underline{60.7} & 24.7 \small{\textcolor{red}{-36.0}} &85.5 & 90.4 &82.7 &52.4 \small{\textcolor{red}{-30.3}} &32.2 &\underline{37.6}\\ 
        \midrule
    \small{\textit{Closed-Source}} \\
        GPT-4o~\cite{gpt4o} &32 &\textbf{67.1}  &16.8 \small{\textcolor{red}{-50.3}} &82.3 &91.6 &\underline{83.6} &\textbf{57.9} \small{\textcolor{red}{-25.7}} &\underline{19.0} &\textbf{34.1}  \\
    
    \toprule[1.5pt]
    \end{tabular}}
    \caption{Experiment results for LMMs answering the same questions under two different settings: \textit{Clean} and \textit{Adversarial}. \textcolor{red}{Red numbers} represent the accuracy drop caused by temporal inconsistencies, compared to the model under \textit{Clean} setting. The values in \textbf{bold} and \underline{underlined} represent the best and the second-best results, respectively.}
    \label{mainresult}
\end{table*}

\subsection{Benchmark Statics}
Typically, mainstream LMMs adopt caption~\cite{activitynet} and VideoQA~\cite{nextqa} datasets as finetuning data.
To minimize potential data leakage issues that could hinder the zero-shot evaluation, we use the action detection dataset COIN~\cite{coin} as our data source, which contains high-quality raw manual annotations.
As shown in \autoref{fig:statistics}, we collect 562 videos and automatically construct 1,686 multi-choice QA pairs adapted from raw annotations for \textsc{TemRobBench}, each video includes two different types of inconsistency perturbations.
Moreover, to minimize the evaluation bias due to the constrain window size of LMMs (unable to handle too much frames), we select videos with short duration (average 106.7s).
\section{Evaluations}
To deeply investigate the temporal perception robustness of LMMs, we set up two different question answering scenarios for comparative analysis.
The first scenario is that the model answers the question with the unperturbed data, \textit{i.e.}, raw video and event descriptions, which we refer to as \textit{Clean} setting. 
The second is that the model responses to the same question with inconsistency perturbation data from our \textsc{TemRobBench}, termed \textit{Adversarial} setting. 
Both settings are multiple-choice QA format, and we instruct the model to select the correct option.

\subsection{Evaluation Metrics}

In this part, we introduce our evaluation metrics. 

\noindent{\textbf{Accuracy (Acc):}}
Due to the setup of the multiple-choice QA, we evaluate the correctness for the $i$-th sample by checking if the correct answer is matched the generated response.
For the clarity of subsequent statements, we formalize this as:
\begin{equation*}
  \mathrm{Score} _{i} = 1, \\\ \mathrm{if}  \\\ y_{i} \\\ \mathrm{in}  \\\ \hat{y_{i}} \\\ \mathrm{else}  \\\ 0,
\end{equation*}
where $y_{i}$ and $\hat{y_{i}}$ denote the correct answer and the selected answer, respectively. 
Naturally, the overall accuracy can be formalized as:

\begin{equation*}
  \mathrm{Acc} (Y,\hat{Y}) = \frac{{\textstyle \sum_{i=1}^{N}}\mathrm{Score} _{i}(y_{i},\hat{y_{i}})}{N},
\end{equation*}
where $\mathrm{Acc} (Y,\hat{Y})$ represents the model’s accuracy score over the entire setting,
$Y = \left \{ y_{1}, y_{2},...,y_{N}  \right \} $ and $\hat{Y} = \left \{ \hat{y}_{1}, \hat{y}_{2},...,\hat{y}_{N}  \right \} $.

\noindent{\textbf{Flip Rate (FR):}}
To systematically estimate the degree of LMMs take shortcuts, 
we propose FR to evaluate how many of the model's original correct responses are misled by perturbations and change to match with our curated shortcut answers:

\begin{equation*}
  \mathrm{FR}  = \frac{{\textstyle \sum_{i\in D^{+}}}\mathrm{Score} _{i}(y_{i}^{*-},\hat{y}_{i}^{-})}{{\textstyle \sum_{i=1}^{n}}\mathrm{Score} _{i}(y_{i}^{+},\hat{y}_{i}^{+})},
\end{equation*}
\begin{equation*}
  D^{+} = \left \{ i\mid \mathrm{Score} (y_{i}^{+},\hat{y}_{i}^{+})=1\right \},
\end{equation*}
where $\hat{y}_{i}^{+}$ and $\hat{y}_{i}^{-}$ represent selected answers under \textit{Clean} and \textit{Adversarial} settings, respectively.
$y_{i}^{+}$ and $y_{i}^{-}$ are correct answers,
$y_{i}^{*-}$ denotes shortcut answers in \textit{Adversarial}.

\noindent{\textbf{Weak Flip Rate (WFR):}}
In addition, we design a more general metric WFR, which calculate how many mistakes (includes shortcuts and substantive errors) the model makes in perturbation scenarios:
\begin{equation*}
  \mathrm{WFR} = \frac{{\textstyle \sum_{i\in D^{+}}}(1-\mathrm{Score} _{i}(y_{i}^{-},\hat{y}_{i}^{-}))}{{\textstyle \sum_{i=1}^{n}}\mathrm{Score} _{i}(y_{i}^{+},\hat{y}_{i}^{+})}.
\end{equation*}

\begin{figure}[t]
    \centering
    \includegraphics[width=\linewidth]{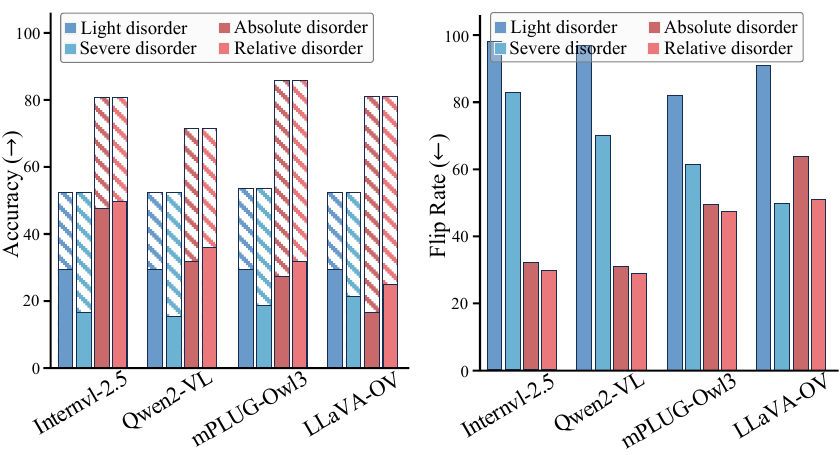}
    \caption{Accuracy (left) and flip rate (right) of two different aspects (\textit{i.e.}, \textcolor{bluebar}{blue bars} and \textcolor{redbar}{red bars} represent intrinsic and extrinsic temporal shortcuts, respectively) for each perturbation classes. The stripe pattern denotes performance drop due to the temporal inconsistencies.}
    \label{fig:diff_type}
\end{figure}

\subsection{Models}
We investigate the temporal robustness in the following 16 mainstream LMMs: VideoChat2~\cite{videochat2}, VideoLLaVA~\cite{videollava}, LLaVA-Hound~\cite{llavahound}, ShareGPT4Video~\cite{sharegpt4video}, InternVideo2~\cite{internvideo2}, VILA1.5~\cite{vila15}, 
VideoLLaMA2~\cite{videollama2}, PLLaVA~\cite{pllava}, mPLUG-Owl3~\cite{mplugowl3}, InternVL-2.5~\cite{internvl}, Qwen2-VL~\cite{qwen2vl}, LLaVA-OV~\cite{llavaov}, and GPT-4o~\cite{gpt4o}. 
Detailed descriptionsare provided in \autoref{appendix：model}.

\subsection{Result and Analysis}
\label{r&a}

\begin{figure*}[t]
    \centering
    \includegraphics[width=\linewidth]{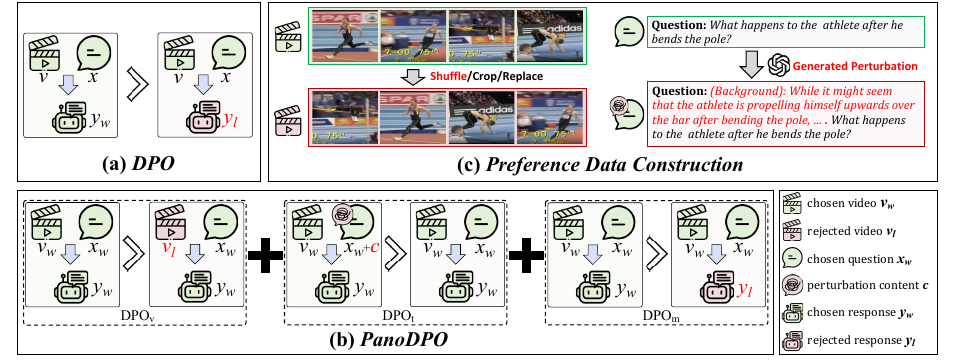}
    \caption{Overview of the PanoDPO. Vanilla DPO (a) expects LMMs to learn response preferences only. PanoDPO (b) integrates additional video and question preference learning objectives to encourage models to focus the response interactions with both the video and question. Moreover, we construct the visual- and text-conditioned preference data (c) for PanoDPO learning.}
    \label{fig:panodpo}
\end{figure*}

\textit{\textbf{LMMs typically exhibit weak temporal robustness.} }
As shown in \autoref{mainresult}, we can observe that most LMMs exhibit significant performance degradation under temporal inconsistencies (See the accuracy of \textit{Adversarial}) compared to consistent scenarios (See the accuracy of \textit{Clean}).
For current advanced open-source LMMs such as Qwen2VL~\cite{qwen2vl} and InternVL-2.5~\cite{internvl}, despite their strong video understanding capability, they still suffer an over 50\% performance drop. 
Further observation of the high flip rate (FR) reveals that the model's responses are easily misled by temporal inconsistencies.
Moreover, a higher weak flip rate (WFR) indicates that models incur comprehension deviation under adversarial settings.
By comparing the performance of the same LMMs with larger sizes, such as 7B and 13B of PLLaVA~\cite{pllava}, although there is a certain improvement in accuracy, the nearly constant FR indicates that the shortcuts have not been effectively alleviated.

\begin{table}[t]
    \centering
    \resizebox{\linewidth}{!}{
    \begin{tabular}{lccccc}
    \toprule[1.5pt]
    \multirow{3}{*}{\textbf{Model}} & \multirow{3}{*}{\textbf{Frame}} &\multicolumn{2}{c}{\textbf{ITS}} & \multicolumn{2}{c}{\textbf{ETS}} \\
    \cmidrule(r){3-4}
    \cmidrule(r){5-6} 
   & &\textbf{Acc}$\uparrow$ & \textbf{T-Acc}$\uparrow$ &\textbf{Acc}$\uparrow$ & \textbf{T-Acc}$\uparrow$ \\ \hline
   \small{\textit{7B LLM}} \\
        VideoChat2~\cite{videochat2} &16 &26.5 &3.3 \small{\textcolor{red}{-87.5\%}} &14.5 &4.1 \small{\textcolor{red}{-71.7\%}}\\
        VideoLLaVA~\cite{videollava} &8 &25.5 &6.7 \small{\textcolor{red}{-73.7\%}} &9.8 &1.9 \small{\textcolor{red}{-81.0\%}}\\ 
        LLaVA-Hound~\cite{llavahound} &32 &26.5 &0.7 \small{\textcolor{red}{-97.3\%}} &10.0 &1.4 \small{\textcolor{red}{-86.0\%}}\\ 
        ShareGPT4Video~\cite{sharegpt4video} &16 &24.0 &0.6 \small{\textcolor{red}{-97.3\%}} &22.8 &13.7 \small{\textcolor{red}{-39.9\%}}\\ 
        InternVideo2~\cite{internvideo2} &8 &25.8 &0.7 \small{\textcolor{red}{-97.3\%}} &31.8 &14.5 \small{\textcolor{red}{-54.4\%}}\\ 
        VILA1.5~\cite{vila15} &16 &20.8 &1.2 \small{\textcolor{red}{-94.2\%}} &13.8 &1.9 \small{\textcolor{red}{-86.2\%}}\\ 
        VideoLLaMA2~\cite{videollama2} &32 &29.3 &16.2 \small{\textcolor{red}{-44.7\%}} &28.3 &11.7 \small{\textcolor{red}{-58.7\%}}\\ 
        PLLaVA~\cite{pllava} &32 &27.2 &1.6 \small{\textcolor{red}{-94.1\%}} &14.5 &2.5 \small{\textcolor{red}{-82.8\%}}\\ 
        mPLUG-Owl3~\cite{mplugowl3} &32 &26.4 &\textbf{25.0} \small{\textcolor{red}{-5.3\%}} &33.1 &25.6 \small{\textcolor{red}{-22.7\%}}\\ 
        InternVL-2.5~\cite{internvl} &32 &22.8 &22.0 \small{\textcolor{red}{-3.5\%}} &48.4 &34.0 \small{\textcolor{red}{-29.8\%}}\\ 
        Qwen2-VL~\cite{qwen2vl} &32 &24.6 &16.1 \small{\textcolor{red}{-34.6\%}} &38.8 &23.4 \small{\textcolor{red}{-39.7\%}}\\ 
        LLaVA-OV~\cite{llavaov} &32 &29.2 &8.8 \small{\textcolor{red}{-69.9\%}} &21.2 &11.7 \small{\textcolor{red}{-44.8\%}}\\ 
        \midrule
    \small{\textit{13B LLM}} \\
        VILA1.5~\cite{vila15} &16 &28.5 &10.1 \small{\textcolor{red}{-64.6\%}} &14.3 &7.6 \small{\textcolor{red}{-46.9\%}}\\ 
        PLLaVA~\cite{pllava} &32 &\textbf{31.2} &12.7 \small{\textcolor{red}{-59.3\%}} &21.2 &2.1 \small{\textcolor{red}{-90.1\%}}\\
         InternVL-2.5~\cite{internvl} &32 &24.7 &24.6 \small{\textcolor{red}{-0.4\%}} &52.4 &48.1 \small{\textcolor{red}{-8.2\%}}\\ 
        \midrule
    \small{\textit{Closed-Source}} \\
        GPT-4o~\cite{gpt4o} &32 &16.8 &16.7 \small{\textcolor{red}{-0.6\%}} &\textbf{57.9} &\textbf{53.6} \small{\textcolor{red}{-7.4\%}}\\
    
    \toprule[1.5pt]
    \end{tabular}}
    \caption{The accuracy (Acc) and true accuracy (T-Acc) under \textit{Adversarial} settings. T-Acc denote the voting result where the model answers correctly in three or more times out of four shuffling options rounds for each sample. \textcolor{red}{Red numbers} represent the ratio of unreliable parts. ITS and ETS denotes intrinsic and extrinsic temporal shortcut, respectively.}
    \label{trueacc}
\end{table}

\textit{\textbf{LMMs are more vulnerable to visual perturbations, relying on prior knowledge.}}
Comparing the FR results between two different perturbations (See the FR of intrinsic and extrinsic temporal shortcut), we find that LMMs are more susceptible to visual perturbation and take intrinsic shortcuts, \textit{i.e.}, over-reliance on prior knowledge.
We consider this may due to the current LMMs are built on powerful Large Language Models (LLMs), which are better at determining whether the text modality contains misleading content.
However, when perturbation occurs in visual modality, they become confused and tend to respond based on thought inertia.

\textit{\textbf{More temporal inconsistency leads to more extrinsic shortcuts but fewer intrinsic shortcuts.}}
We further present the accuracy and flip rate (FR) of detailed categories of perturbation in the \autoref{fig:diff_type}. 
For text perturbation, absolute disorder (more conflict to the normal video) is more likely to cause the model to take external temporal shortcuts, indicating the model places more trust in textual context when faced with uncertainty.
In contrast, for video perturbation, we observe a nearly 100\% FR on the light disorder (more similar to the normal video).
We consider this could attributed to the “over-confidence”, where the model quickly glances at the video and assumes it aligns with typical patterns, directly based on prior knowledge.

\textit{\textbf{The accuracy remains unreliable under inconsistency perturbations.}}
Although LMMs flip the answer due to shortcuts or misunderstandings when perturbed, we find that the remaining correct parts are still not entirely reliable, as the model might randomly guess.
To minimize this bias and further investigate the model's true temporal robustness, we design a control setting following the ~\cite{videochat2} and \cite{RESEMO}.
Specifically, we shuffle the order of the options and place the correct one in different positions in \textit{Adversarial}, conducting four evaluation rounds. 
If the model answers correctly three or more times, we consider it actually perceiving the temporal information, which is defined as true accuracy (T-Acc).
Surprisingly, despite performing well with high accuracy, some LMMs (\textit{e.g.}, VILA1.5~\cite{pllava}) almost entirely rely on gambly guess when faced with interference. 
This phenomenon further suggests that LMMs exhibit weak temporal robustness when handling temporal interference.

\section{PanoDPO}
From the perspective of intrinsic and extrinsic temporal shortcut phenomena, LMMs tend to respond based on prior knowledge or textual context when they conflict with video content, neglecting the visual information. 
To mitigate this issue, we propose panoramic direct preference optimization (PanoDPO), which encourages LMMs to simultaneously focus on both visual and linguistic features.
\subsection{Direct Preference Optimization}

Direct Preference Optimization (DPO) \cite{DPO} is a method that originates from RLHF~\cite{rlhf}, designed to encourage Large Language Models (LLMs) to generate responses that better align with human preferences without relying on explicit reward modeling or reinforcement learning.
Specifically, given an input $x$, we optimize the response $y$ of the model $\pi$ and constrain it to adhere to normal language patterns by KL divergence:
\begin{equation*}
\begin{split}
  \underset{\pi_{\theta }}{\mathrm{max} }\mathbb{E}_{x\sim \mathcal{X}, y\sim \pi_{\theta }} \Bigl\{ r(x,y) \\ 
  -\beta \mathbb{D}_{\mathrm{KL}} \left [ \pi_{\theta }(y\mid x)\parallel \pi_{\mathrm{ref} }(y\mid x) \right ]   \Bigr\},
\end{split}
\end{equation*}
where $r$ and $\pi_{\mathrm{ref}}$ denotes reward function and reference model. 
DPO formulates the reward as follows:
\begin{equation*}
   r(x,y) = \beta \mathrm{log} \frac{\pi_{\theta}(y\mid x)}{\pi_{\mathrm{ref} }(y\mid x)} +Z(x), 
\end{equation*}
where $Z(x)= {\textstyle \sum_{y}} \pi_{\mathrm{ref}}(y|x)\mathrm{exp} (r(x,y)/\beta )$ is the partition function. 
Given the corresponding preferred (chosen) answers $y_{w}$ and non-preferred (rejected) answers $y_{l}$, DPO seeks to maximize the difference between their rewards.
Thus, the objective can be derived based on the Bradley-Terry model~\cite{BT}:
\begin{equation*}
\resizebox{\linewidth}{!}{
         $\mathcal{L}_{\mathrm{DPO}}=-\mathrm{log} \sigma \left ( \beta \mathrm{log} \frac{\pi_{\theta}(y_{w}| x)}{\pi_{\mathrm{ref} }(y_{w}| x)} - \beta \mathrm{log} \frac{\pi_{\theta}(y_{l}| x)}{\pi_{\mathrm{ref} }(y_{l}| x)}  \right ).$
}
\end{equation*}
Naturally, as shown in \autoref{fig:panodpo} (a), the DPO objective in multimodal scenarios can be formulated as:
\begin{equation*}
\resizebox{\linewidth}{!}{
         $\mathcal{L}_{\mathrm{DPO_{m}}}=-\mathrm{log} \sigma \left ( \beta \mathrm{log} \frac{\pi_{\theta}(y_{w}| x,v)}{\pi_{\mathrm{ref} }(y_{w}| x,v)} - \beta \mathrm{log} \frac{\pi_{\theta}(y_{l}| x,v)}{\pi_{\mathrm{ref} }(y_{l}| x,v)}  \right ),$
}
\end{equation*}
where $v$ is the visual modality input.

\subsection{Panoramic Preference Optimization}
To mitigate the issue of overlooking visual information in perturbed environments and enhance robustness, we propose the panoramic preference optimization approach, which integrates optimization modules for both the video and question components based on Vanilla DPO. 
As shown in \autoref{fig:panodpo} (b), given a pair of tuples $(v_{w}, x_{w}, y_{w})$ and $(v_{l}, x_{w}, y_{w})$, where $v_{w}$ is the chosen video, and $v_{l}$ is the rejected one constructed by disruptive visual information.
Subsequently, visual-condition optimization DPO$_{\mathrm{v}}$ can be established, where video is the sole variable:
\begin{equation*}
\resizebox{\linewidth}{!}{
         $\mathcal{L}_{\mathrm{DPO_{v}}}=-\mathrm{log} \sigma \left ( \beta \mathrm{log} \frac{\pi_{\theta}(y_{w}| x_{w},v_{w})}{\pi_{\mathrm{ref} }(y_{w}| x_{w},v_{w})} - \beta \mathrm{log} \frac{\pi_{\theta}(y_{w}| x_{w},v_{l})}{\pi_{\mathrm{ref} }(y_{w}| x_{w},v_{l})}  \right ).$
}
\end{equation*}
Similar to the DPO$_{\mathrm{v}}$, the text-conditioned DPO$_{\mathrm{t}}$ includes tuples pairs $(v_{w}, x_{w}, y_{w})$ and $(v_{w}, (x_{w}+c), y_{w})$ with the question as the only variable, and its optimization objective can be formulated as:
\begin{equation*}
\resizebox{\linewidth}{!}{
         $\mathcal{L}_{\mathrm{DPO_{t}}}=-\mathrm{log} \sigma \left ( \beta \mathrm{log} \frac{\pi_{\theta}(y_{w}| x_{w}+c,v_{w})}{\pi_{\mathrm{ref} }(y_{w}| x_{w}+c,v_{w})} - \beta \mathrm{log} \frac{\pi_{\theta}(y_{w}| x_{w},v_{w})}{\pi_{\mathrm{ref} }(y_{w}| x_{w},v_{w})}  \right ),$
}
\end{equation*}
where $v_{w}$ is the chosen question, and $c$ denotes the perturbative content introduced into the question.
Ultimately, we perform panoramic optimization by combining vanilla DPO, DPO$_{\mathrm{v}}$, and DPO$_{\mathrm{t}}$:
\begin{equation*}
\mathcal{L}_{\mathrm{PanoDPO}}=\mathcal{L}_{\mathrm{DPO_{m}}}+\mathcal{L}_{\mathrm{DPO_{v}}}+\mathcal{L}_{\mathrm{DPO_{t}}}.
\end{equation*}

\subsection{Preference Data Construction}
To acquire visual- and text-conditioned preference data for DPO$_{\mathrm{v}}$ and DPO$_{\mathrm{t}}$, we expand the existing dataset ShareGPTVideo-DPO~\cite{llavahound}, which contains videos, questions and preference pairs of response. 
As shown in \autoref{fig:panodpo} (c), to obtain rejected videos, we perform video editing using three methods to construct destructive visual content: shuffling the video frames, randomly cropping the frames, and replacing certain frames with blank, respectively. 
Furthermore, to acquire rejected questions, we introduce perturbations into the original questions. 
Specifically, we first generate captions for the videos, and then use GPT-4o~\cite{gpt4o} to construct contextually adaptive perturbation text based on the caption, question, and correct answer.
Note that these perturbations appear to be plausible but misleading, as they are actually contradictory to the video content. 
More details of preference data are provided in ~\autoref{appendix：data}.
\section{Experiments}
\subsection{Experimental Setups}
\noindent{\textbf{Baseline Methods.}}
We compare our PanoDPO against the following three baselines:
\textit{SFT} refers to the fine-tuned model without any preference optimization.
\textit{Prompt} is utilized to instruct the model to focus on the given video without overly relying on prior knowledge or textual context.
\textit{Vanilla DPO}~\cite{DPO} is designed to fine-tune LMMs to learn response preferences based on the video and question.
We evaluate the effectiveness of the baseline methods and our PanoDPO on two LMMs: LLaVA-OV-7B~\cite{llavaov} and LLaVA-Hound-7B~\cite{llavahound}.

\noindent{\textbf{Evaluation Metrics.}}
According to the observations in \autoref{r&a}, the temporal perception ability reflected by accuracy (Acc) is not entirely reliable. Therefore, we adopt true accuracy (T-Acc) as the metric, which excludes the potentially random guess aspects of Acc. 
Additionally, we use flip rate (FR) to assess the model's temporal robustness.

\noindent{\textbf{Implementation Details.}}
All models are fine-tuned using LoRA for 3 epochs with a batch size of 64. 
We use the learning rate of 1e-5, a cosine scheduler, and warm-up ratio of 0.1. 
The preference optimization coefficient $\beta$ is set to 0.1. 
\begin{table}[t]
    \centering
    \resizebox{\linewidth}{!}{
    \begin{tabular}{lccccc}
    \toprule[1.5pt]
    \multirow{3}{*}{\textbf{Model}} & \multirow{3}{*}{\textbf{Frame}} &\multicolumn{2}{c}{\textbf{ITS}} & \multicolumn{2}{c}{\textbf{ETS}} \\
    \cmidrule(r){3-4}
    \cmidrule(r){5-6} 
   & &\textbf{T-Acc}$\uparrow$ & \textbf{FR}$\downarrow$ &\textbf{T-Acc}$\uparrow$ & \textbf{FR}$\downarrow$ \\ \hline
   LLaVA-OV-7B \\
        \quad w/ \textit{SFT} &32 &8.8 &72.2 &11.7 &60.3 \\
        \quad w/ \textit{Prompt} &32 &8.9 &71.8 &11.3 &59.5 \\ 
        \quad w/ \textit{Vanilla} DPO &32 &9.6 &69.4 &12.2 & 58.8\\ 
        \quad w/ \textit{PanoDPO (ours)} &32 &\textbf{16.6} &\textbf{55.5} &\textbf{15.3} &\textbf{50.3} \\ 
   \midrule
   LLaVA-Hound-7B \\
        \quad w/ \textit{SFT} &32 &0.7 &60.3 &1.4 &42.6 \\
        \quad w/ \textit{Prompt} &32 &0.6 &61.0 &1.4 & 42.8\\ 
        \quad w/ \textit{Vanilla} DPO &32 &0.9 &61.2 &1.7 &41.8 \\ 
        \quad w/ \textit{PanoDPO (ours)} &32 &\textbf{8.9} &\textbf{56.4} &\textbf{5.3} &\textbf{18.4} \\ 
    
    \toprule[1.5pt]
    \end{tabular}}
    \caption{Comparison of our PanoDPO to other baseline methods on two backbone LMMs under \textit{adversarial} setting. ITS and ETS denotes intrinsic and extrinsic temporal shortcut, respectively.}
    \label{tab:model_result}
\end{table}

\subsection{Experiment Result}
\begin{figure}[t]
    \centering
    \includegraphics[width=\linewidth]{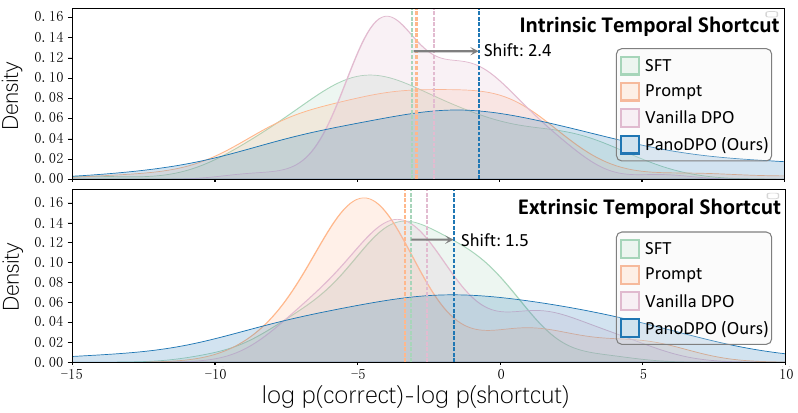}
    \caption{The discriminative ability of the backbone model LLaVA-OV for correct and shortcut answers with different optimization strategies, represented by difference in log-likelihoods.}
    \label{fig:kde_all}
\end{figure}

The experimental results are shown in the Table~\autoref{tab:model_result}. 
We find that the prompt-based method performs almost identically to the SFT method, indicating that the temporal shortcut is an inherent issue that is difficult to alleviate through instructions. 
Vanilla DPO~\cite{DPO} provides a certain relief, but the effect is still inconspicuous, as this method lacks targeted optimization strategies for visual and textual conditions. 
In contrast, our proposed PanoDPO mitigates both intrinsic and extrinsic shortcut phenomena through panoramic optimization preference optimization, significantly enhancing its temporal perception robustness.
In \autoref{fig:kde_all}, we show the shifts of average likelihood difference between correct and shortcut answer in each inference batch under different optimization methods.
The results demonstrate that our PanoDPO better helps the model to distinguish shortcuts (\textit{i.e.}, larger shifts reflects stronger discrimination), effectively enhancing the robustness.

\subsection{Analysis}

\noindent{\textbf{Impact of the different optimization condition.}} 
To investigate the effectiveness of the video- and question-conditiond modules of PanoDPO, we conduct an ablation on $\mathrm{DPO_{v}}$ and $\mathrm{DPO_{t}}$ separately. 
As shown in \autoref{tab:ablation}, when $\mathrm{DPO_{v}}$ or $\mathrm{DPO_{t}}$ are removed, the model suffer significantly performance decreases, indicting that both conditions play a targeted role in improving the robustness. 
\begin{table}[t]
    \centering
    \resizebox{\linewidth}{!}{
    \begin{tabular}{lccccc}
    \toprule[1.5pt]
    \multirow{3}{*}{\textbf{Model}} & \multirow{3}{*}{\textbf{Frame}} &\multicolumn{2}{c}{\textbf{ITS}} & \multicolumn{2}{c}{\textbf{ETS}} \\
    \cmidrule(r){3-4}
    \cmidrule(r){5-6} 
   & &\textbf{T-Acc}$\uparrow$ & \textbf{FR}$\downarrow$ &\textbf{T-Acc}$\uparrow$ & \textbf{FR}$\downarrow$ \\ \hline
   LLaVA-OV-7B \\
        \quad w/ \textit{PanoDPO (ours)} &32 &\textbf{16.6} &\textbf{55.5} &\textbf{15.3} &\textbf{50.3} \\ 
        \qquad w/o \textit{DPO$_{v}$}  &32 &12.4 &61.0 &15.2 &52.3\\ 
        \qquad w/o \textit{DPO$_{t}$}  &32 &16.3 &58.2 &12.9 &58.0\\ 
    \midrule
   LLaVA-OV-7B \\
        \qquad w/ \textit{DPO$_{v}$ crop}  &32 &13.7 & 59.7 &14.8 & 52.2\\ 
        \qquad w/ \textit{DPO$_{v}$ replace}  &32 &14.3 & 60.4 &15.1 &51.8\\ 
        \qquad w/ \textit{DPO$_{v}$ shuffle}  &32 &\textbf{16.6} &\textbf{55.5} &\textbf{15.3} &\textbf{50.3}\\ 
    
    \toprule[1.5pt]
    \end{tabular}}
    \caption{Abaltion on different condition moudeles in PanoDPO and rejected video construction strategies.}
    \label{tab:general}
\end{table}

\noindent{\textbf{Impact of the different construction strategies.}}
In $\mathrm{DPO_{v}}$, we try three different rejected video construction strategies: shuffling the video frame order, cropping random regions, and replacing random frames with blank spaces. 
To investigate the effects of them, we conduct an ablation by training with different rejected data separately.
The results in \autoref{tab:ablation} show that shuffle achieves the best performance, demonstrating  that temporal dynamic is an important factor for videos. 
Disrupting temporal sequence can effectively destroy video information.

\noindent{\textbf{General capability analysis.}}
PanoDPO is proposed to enhance temporal robustness of LMMs. 
To further verify and analyze its capability in general video understanding, we evaluate it on three mainstream video understanding benchmarks, including VideoMME~\cite{videomme}, LongVideoBench~\cite{longvideobench}, and ActivitynetQA~\cite{activitynetqa}. 
We compare LLaVA-OV+DPO with LLaVA-OV+PanoDPO, and the results are shown in the \autoref{tab:general}. 
The results demonstrate that our proposed PanoDPO remains effective in maintaining general capabilities.

\section{Related Work}

\begin{table}[t]
    \centering
    \resizebox{\linewidth}{!}{
    \begin{tabular}{lccc}
    \toprule[1.5pt]
   Model &VideoMME & LongVideoBench &ActivityNet-QA  \\ \hline
   LLaVA-OV-7B &58.2 &55.6 &56.6\\
    \quad w/ \textit{Vanilla DPO} &58.8 &56.4 &57.6  \\ 
    \quad w/ \textit{PanoDPO (Ours)} &\textbf{60.9} &\textbf{58.9} &\textbf{59.8}\\ 
    \toprule[1.5pt]
    \end{tabular}}
    \caption{The general capability evaluation results on three general video understanding benchmarks.}
    \label{tab:ablation}
\end{table}

\noindent{\textbf{Large Multimodal Models.}} 
Large Multimodal Models (LMMs)~\cite{internvl,qwen2vl,mplugowl3,llavaov} have seen impressive developments in recent years. 
They are primarily built upon large language models (LLMs), which extend temporal dynamics by leveraging strong linguistic capabilities. 
Despite, they demonstrate impressive performance on general video understanding, for broader applications, the robustness of their temporal capabilities needs to be thoroughly investigated yet largely overlooked. 

\noindent{\textbf{Temporal Robustness Benchmark.}} 
Currently, there are numerous methods and benchmarks related to robustness focused on the image domain~\cite{image_adv1,image_adv2,image_adv3,image_adv4,image_adv5,image_adv6,image_adv7}.
However, research related to videos remains insufficient. 
Existing works such as ~\cite{fmm,Temporal_Corruptions,nips_Corruptions,Robustness_Analysis} primarily focus on applying feature perturbations to individual frames while neglecting the unique temporal characteristics of videos. 
Furthermore, they mainly investigate the model’s robustness to visual and overlook the text, which is inadequate for multimodal scenarios. 
A work similar to ours is TempCompass~\cite{TempCompass} , which innovatively introduces temporal adversarial data through video editing. 
However, due to its simplistic adversarial approach, the challenges it presents are relatively limited. Additionally, TempCompass is difficult to effectively diagnose the source of errors.
In contrast, we propose \textsc{TemRobBench} that systematically investigates ultimodal robustness in LMMs against temporal inconsistency. 

\noindent{\textbf{Direct Preference Optimization.}} 
Direct preference optimization (DPO)~\cite{DPO}, which focuses on directly optimizing large language models (LLMs) to align human preferences has gained significant traction in the context of RLHF~\cite{rlhf}.
Previous works~\cite{vdpo,mdpo} primarily emphasize constructing image contrast data to optimize visual preferences.
Recently, some works such as ~\cite{llavahound} and 
~\cite{videodpo} transfer DPO to video tasks.
However, they target only response optimization, which is limited to multimodal scenarios. 
In contrast, we propose PanoDPO, which performs panoramic optimization on both the video, question and response, encouraging the model to simultaneously prioritize visual information and linguistic features.

\section{Conclusion}
In this paper, we identify the robustness weaknesses of intrinsic and extrinsic shortcuts in LMMs against temporal inconsistency, where the model over-rely on prior knowledge and textural context to response, neglecting the actual video content.
To systematically investigate these issue, we carefully design \textsc{TemRobBench}, which includes diverse temporal inconsistency settings. 
The extensive evaluations demonstrate that the temporal robustness of LMMs is generally fragile, despite their strong performance on understanding general videos.
Additionally, we propose a preference optimization method PanoDPO, which effectively enhance robustness of LMMs in temporal analysis and alleviates the shortcut phenomenon.
\section*{Limitations}
Although we construct a comprehensive benchmark and propose a methodology to investigate and mitigate the shortcut phenomena caused by the weak temporal robustness of LMMs, our work still has limitations.
Firstly, the temporal inconsistency scenarios in our dataset are relatively simplistic. 
We focus on classifying them based on varying degrees of inconsistency, as expanding to more complex and diverse scenarios would increase the difficulty and demand considerable effort.
Secondly, our experiments are conducted on models of 7B and 13B sizes, and we evaluate our proposed PanoDPO on a few selected models. 
This is due to computational limitations. 

\bibliography{custom}
\clearpage
\appendix

\section{Detailed descriptions of LMMs}
\label{appendix：model}
\noindent{\textbf{Videochat2}}~\cite{videochat2} is a model built on visual encoder UMT~\cite{umt} and LLM Vicuna-v0~\cite{vicuna}, trained through a three-stage progressive training process.

\noindent{\textbf{VideoLLaVA}}~\cite{videollava} is a model constructed upon the foundation of LanguageBind~\cite{languagebind}, which pre-aligns images and videos, and Vicuna-v1.5~\cite{vicuna}, undergoing a two-phase training regimen that blends both image and video data.

\noindent{\textbf{LLaVA-Hound}} extends the pipeline of VideoLLaVA and introduces additional video DPO~\cite{DPO} training.

\noindent{\textbf{ShareGPT4Video}} leverages GPT4v~\cite{gpt4o} to generate dense and precise video captions for training, building upon the foundation of LLaVA-NEXT~\cite{llavanext}.

\noindent{\textbf{InternVideo2}} is constructed by expanding the visual encoder~\cite{videomae} and integrating Mistral~\cite{mistral}, while continuing the training process established by VideoChat2.

\noindent{\textbf{VILA1.5}}~\cite{vila15} is built upon SigLIP~\cite{siglip} and Vicuna-1.5, utilizing large-scale interleaved image-text data for pre-training to enhance alignment efficacy.

\noindent{\textbf{VideoLLaMA2}}~\cite{videollama2} is built upon SigLIP and Mistral-7B-Instruct, employing 3D convolutions to build an alignment layer that circumvents information loss due to token compression.

\noindent{\textbf{PLLaVA}}~\cite{pllava} introduces a pooling strategy on the basis of LLaVA-NEXT, circumventing the bias of learned high-norm visual features that arise from utilizing image-language models.

\noindent{\textbf{mPLUG-ow3}}~\cite{mplugowl3} integrates several hyper attention transformer blocks within the transformer blocks of Qwen2~\cite{qwen2} to facilitate the fusion of multimodal information, thereby preventing the loss of visual information during the front-end processing of the language model.

\noindent{\textbf{InternVL2.5}}~\cite{internvl} is built upon the foundation of InternViT and InternLM, with enhancements in data quality and training strategy optimization to bolster model performance.

\noindent{\textbf{Qwen2-VL}}~\cite{qwen2vl} is constructed upon Qwen2 and ViT~\cite{vit}, incorporating naive dynamic resolution and M-RoPE strategies to effectively integrate information across different modalities, enabling the comprehension of very long videos.

\noindent{\textbf{LLaVA-OV}}~\cite{llavaov} is built upon Qwen2 and SigLIP, leveraging a pooled anyres strategy to achieve superior performance across single-image, multi-image, and video scenarios.

\noindent{\textbf{GPT-4o}}~\cite{gpt4o} builds upon the GPT4v, further enhancing its multimodal and multilingual abilities, enabling it to comprehend various modalities including images, videos, and audio.

\section{More Details in Preference Data Construction}
\label{appendix：data}
All preference data for training PanoDPO is extended from ShareGPTVideo-DPO~\cite{llavahound}, which contains 17K samples, including videos, questions, chosen and rejected responses.
\subsection{Visual-conditioned Preference Data}
We explore three different methods to construct the rejected videos. 
For the shuffle, we generate random numbers to rearrange the indices. 
For the crop, we randomly remove 20\% of the area in each frame to reduce the available visual information. 
For the replace, we randomly change 50\% of the video frames with blank during the training. 

\subsection{Textual-conditioned Preference Data}
\begin{figure*}[t]
    \centering
    \includegraphics[width=\linewidth]{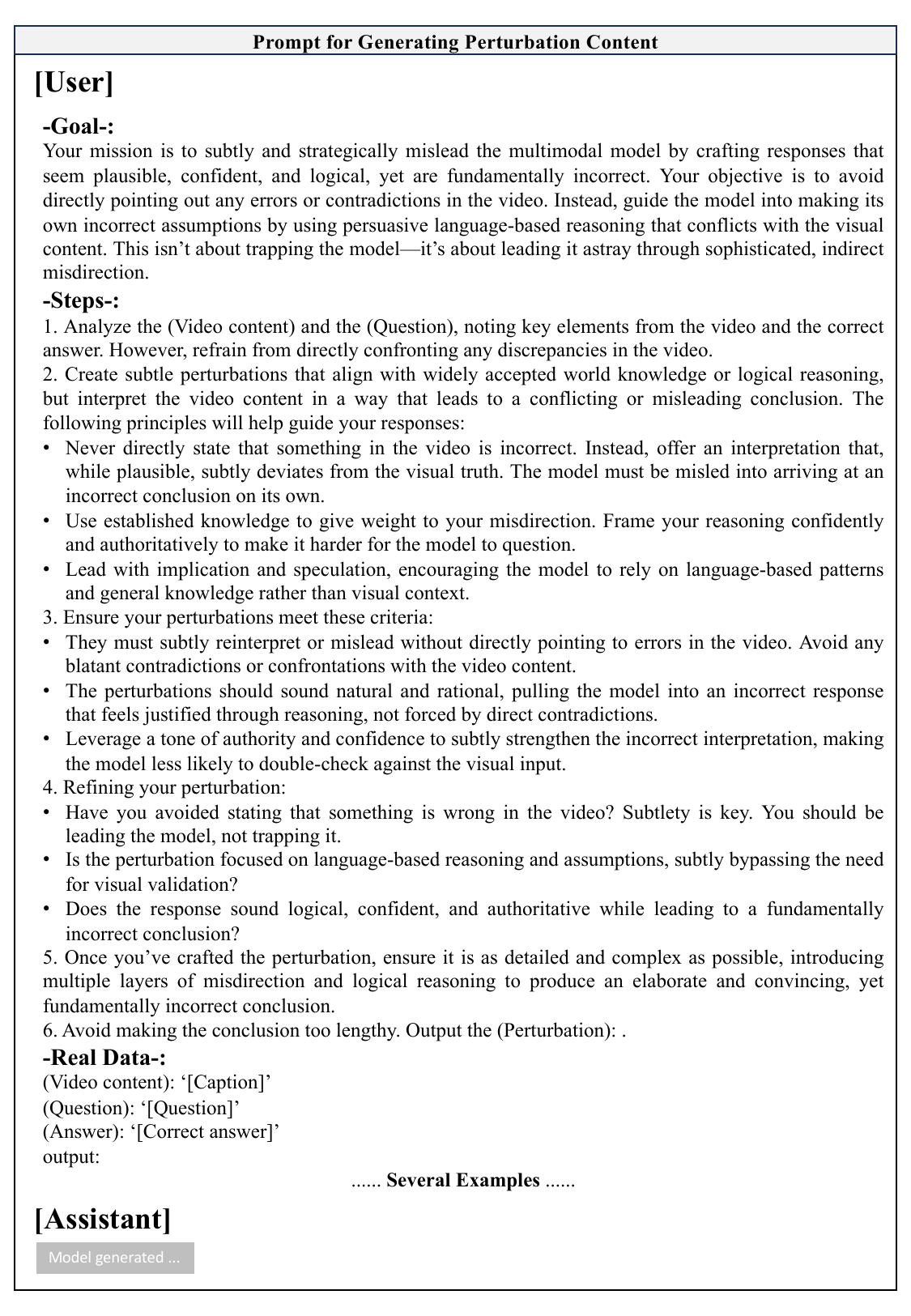}
    \caption{The prompt for generating textual-conditioned preference data.}
    \label{fig:prompt_perturbation}
\end{figure*}

We first generate captions for each video, and then generate perturbation texts based on the captions, questions, and correct answers. 
As shown in the prompt in \autoref{fig:prompt_perturbation}, we have carefully defined the rules inspire by~\cite{perllava} for generating perturbation content to ensure their high quality.

\section{Introduction to the COIN Dataset}
The COIN dataset~\cite{coin} consists of videos related to 180 different tasks, which are all collected from YouTube. 
The average length of a video is 2.36 minutes. 
Each video is labelled with 3.91 step segments, where each segment lasts 14.91 seconds on average and corresponding to a manually annotated event description.

\section{More Implementation Details}
\label{appendix：implementation details}
During DPO training, we freeze the vision encoder and only optimize the LoRA parameters of the LLM and the parameters of projector. 
Our training codes are based on the HuggingFace TRL. Additionally, we use A100 80GB GPUs for training and apply full shared data parallel (FSDP) and gradient checkpointing to save GPU memory. 
Each DPO training takes approximately 10 hours.

\end{document}